\documentclass[letterpaper, 10 pt, conference]{ieeeconf}  % Comment this line out if you need a4paper

\IEEEoverridecommandlockouts                              % This command is only needed if 
\usepackage{times}
\usepackage{latexsym}
\usepackage{multicol}
\usepackage{algorithm}
\usepackage{algorithmicx}
\usepackage{algpseudocode}
\usepackage{graphicx}
\graphicspath{{images/}}
\usepackage{subfigure}
\usepackage{float}

\usepackage{booktabs}
\usepackage{multirow}
\usepackage{array}
\usepackage{amsmath}
\usepackage{microtype}
\usepackage{xcolor}
\usepackage{hhline}

\usepackage{hyperref}

\usepackage{amssymb}

\floatname{algorithm}{Class} % Change the caption name to "Class"

\title{\LARGE \bf
SPACE: A Python-based Simulator for Evaluating Decentralized Multi-Robot Task Allocation Algorithms
}

\author{Inmo Jang$^{*}$% <-this % stops a space
\thanks{The authors is with Department of Autonomous Vehicle Engineering, Korea Aerospace University, Goyang, Gyeonggi, 10540, South Korea.
         }%
\thanks{$^{*}$\textit{corresponding author:} {\tt\small inmo.jang@kau.ac.kr}}%
}

\begin{document}
\maketitle{}
\thispagestyle{empty}
\pagestyle{empty}

%%%%%%%%%%%%%%%%%%%%%%%%%%%%%%%%%%%%%%%%%%%%%%%%%%%%%%%%%%%%%%%%%%%%%%%%%%%%%%%%
\begin{abstract}

Swarm robotics explores the coordination of multiple robots to achieve collective goals, with \emph{collective decision-making} being a central focus. 
This process involves decentralized robots autonomously making local decisions and communicating them, which influences the overall emergent behavior.
Testing such decentralized algorithms in real-world scenarios with hundreds or more robots is often impractical, underscoring the need for effective simulation tools. 
We propose \emph{SPACE} (Swarm Planning and Control Evaluation), a Python-based simulator designed to support the research, evaluation, and comparison of decentralized Multi-Robot Task Allocation (MRTA) algorithms. 
SPACE streamlines core algorithmic development by allowing users to implement decision-making algorithms as Python plugins, easily construct agent behavior trees via an intuitive GUI, and leverage built-in support for inter-agent communication and local task awareness. 
To demonstrate its practical utility, we implement and evaluate CBBA and GRAPE within the simulator, comparing their performance across different metrics, particularly in scenarios with dynamically introduced tasks.
This evaluation shows the usefulness of SPACE in conducting rigorous and standardized comparisons of MRTA algorithms, helping to support future research in the field.
\end{abstract}

%%%%%%%%%%%%%%%%%%%%%%%%%%%%%%%%%%%%%%%%%%%%%%%%%%%%%%%%%%%%%%%%%%%%%%%%%%%%%%%%
\section{Introduction}

% Swarm Robotics 
Swarm robotics is a field that studies the coordination of multiple robots to perform tasks collectively, offering promising potential for future technological advancements. 
% Collective Decision-making process in Swarm Robotics
A unique characteristic of control strategies in swarm robotics  is their high-level layer called the \emph{collective decision-making} process  \cite{Valentini2015}, where each robot evaluates available options, selects one of them, and then communicates this local decision to its neighboring agents.
% Emergent behaviour is important in this process
This process, coupled with the large number of robots typically involved in swarm robotics, leads to significant inter-robot interactions that influence the overall emergent behavior. 
Analyzing these interactions and understanding how to achieve desired emergent behaviors is a central research focus \cite{Dorigo2014}. 
% Importance of Simulation
While real-world experimentation with dozens of robots is feasible, testing decentralized algorithms with hundreds or more robots is impractical in an academic setting, emphasizing the crucial role of simulation in this field \cite{Cindy2022}.

% Traditional Simulation in MRTA Research
%% Non Standardized
Several simulators exist for robotics, although few are specifically tailored for studying collective decision-making algorithms in swarm robotics (see Section \ref{sec_related_work_simulator} for a review of existing simulators). 
Evaluating high-level decision-making algorithms, such as \emph{Multi-Robot Task Allocation (MRTA)} problems, does not necessarily require highly powerful computing resources. 
Decentralized algorithms involve each agent performing local decision-making based on local information, keeping computational demands relatively low. 
Moreover, MRTA algorithms often do not necessitate high-fidelity individual-level robot physics, as decision-making occurs in an abstract layer. % (e.g., utility and cost for each robot). 
Consequently, researchers often implement their own simulations using MATLAB or Python on standard computers to evaluate their proposed algorithms. 
However, this approach lacks standardization, leading to varied and inconsistent evaluation methods. 
% Therefore, there is a need for a simulator focused on high-level decision-making algorithms, allowing researchers to concentrate on the core aspects of their work and minimise redundant implementation efforts.

\begin{figure}[t]
\centering
\includegraphics[width=0.99
\linewidth]{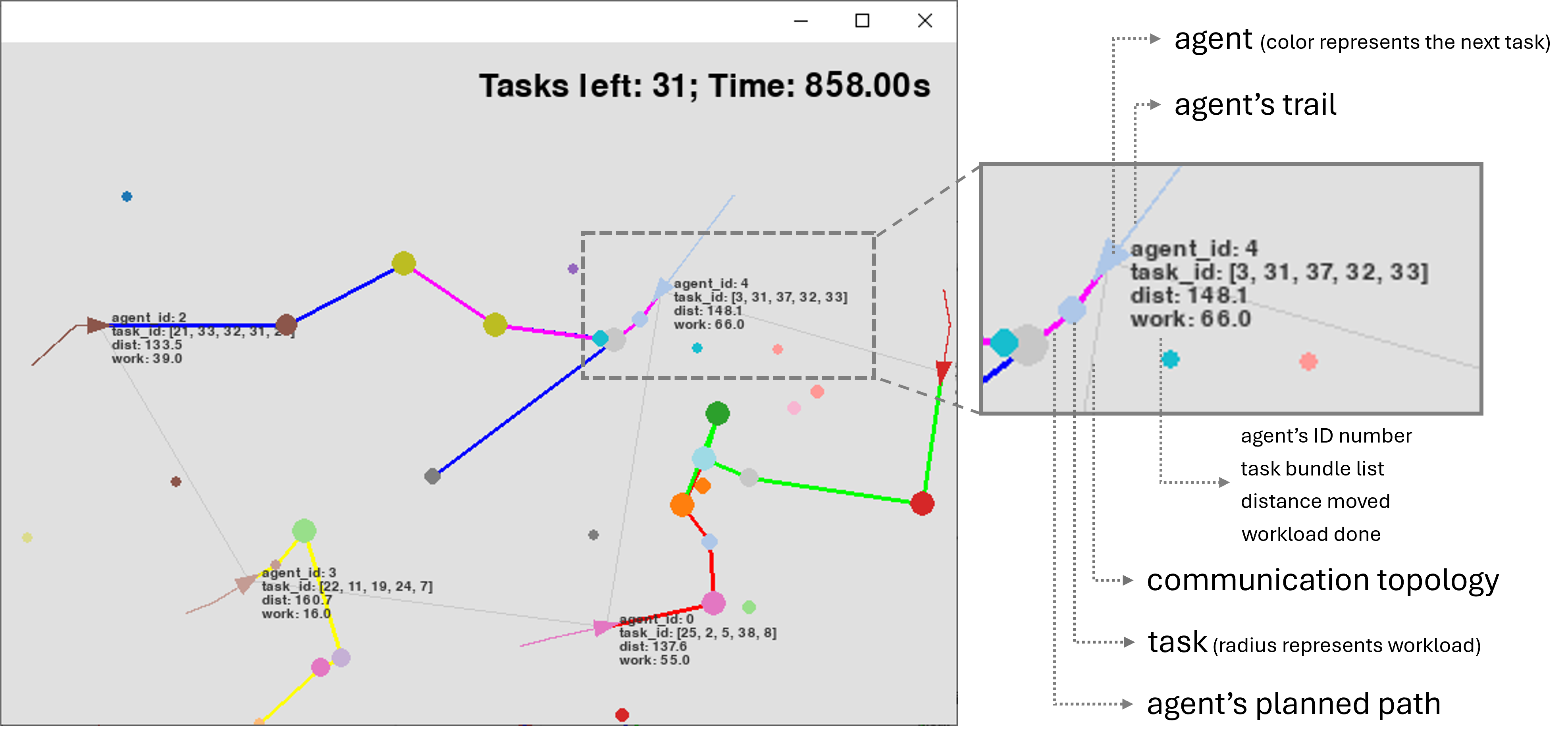}
\caption{SPACE simulator visualization example ($n_a=5; n_t=40$) with the CBBA plugin in Section \ref{sec_implementation}}
\label{fig_example}
\end{figure}

%% Static Scenario 
Evaluating MRTA algorithms in dynamic situations is also valuable but requires additional effort in simulation setup. 
Many MRTA studies primarily focus on static scenarios, where tasks are defined before a mission begins, and agents calculate the assignments \cite{Chopra2017, Choi2009}. 
Although there is some exploration of dynamic task generation scenarios, most research does not simulate environmental changes resulting from agent movement or task progression. 
Instead, these evaluations often consider cases where new tasks are introduced after an agreed assignment and measure the computational time required for re-convergence \cite{Johnson2011, Jang2018}. 
Recently, analysis of sequential decision-making in dynamic task generation scenarios have emerged \cite{Park2021, Choudhury2022}. Nevertheless, researchers still need to implement these features in their own simulators for thorough evaluation.

% Purpose of this work: MRTA Simulator
This study introduces \emph{SPACE} (Swarm Planning and Control Evaluation), a simulator designed to facilitate the research, evaluation, and comparison of decentralized MRTA algorithms with minimal coding effort. 
Users only need to develop their own decision-making algorithm as a Python plugin and configure agents to use it through YAML settings.
SPACE comes equipped with built-in features for local communication and local situational awareness, and supports dynamic task generation. 
By implementing the core of the agent controller based on \emph{behavior trees} (BTs) \cite{Colledanchise2021, Ghzouli2023}, the simulator facilitates easy development of necessary agent-level behaviors surrounding the decision-making algorithm.
Thanks to this feature, SPACE facilitates effective comparisons of algorithms across different MRTA categories \cite{Gerkey2004}, for example, evaluating ST-MR algorithms like GRAPE \cite{Jang2018} against MT-SR algorithms like CBBA \cite{Choi2009} in a specific mission scenario.

\section{Related Work}

\subsection{Simulators for Swarm Robotics}\label{sec_related_work_simulator}

% ### Comparative Analysis of Simulators for High-Level Collective Decision-Making Algorithms

Each simulator has distinct design goals, and this fact makes direct comparisons not straightforward \cite{Erez2015}.
Despite the challenge, this section aims to review the features of simulators that have been ever updated in the past five years \cite{Cindy2022}, specifically focusing on {Gazebo}, {Webots}, {CoppeliaSim}, {ARGoS}, and {Stage}, from the perspective of high-level collective decision-making algorithm research.

% ### High-Fidelity Simulators: Webots, Gazebo, and V-REP
\emph{Gazebo}, \emph{Webots}, \emph{CoppeliaSim} are well-known for their high-fidelity simulations in the robotics domain. 
They provide precise physical modeling, creating realistic environments for testing and validating robotic algorithms. 
However, their focus on high-fidelity makes real-time simulation of even dozens of robots challenging \cite{Pinciroli2012}, limiting their effectiveness for large-scale swarm robotics studies.

% ### ARGoS: A Balance Between Performance and Realism
\emph{ARGoS} \cite{Pinciroli2012} is a simulator that balances realism and performance, efficiently managing larger robot swarms through its modular architecture that accommodate various simulation requirements. 
It supports agent-level programming, as well as local communication and sensing. 
However, despite its use in evaluating multi-robot task allocation algorithms \cite{Kang2024}, the need for C++ implementation to set up simulations with specific MRTA features can be challenging for researchers who prefer higher-level languages for rapid prototyping, quick testing, and comparative analysis.

% ### Stage: Lightweight but Limited
\emph{Stage} \cite{Vaughan2008} is known for its lightweight nature, making it one of the simplest simulators available for multi-robot systems. 
Although it natively supports agent local sensing, it lacks support for local communication. Additionally, its last update was nearly five years ago, and its documentation and tutorials are not comprehensive enough.

% ### SPACE: Tailored for MRTA and Ease of Use

Our work was inspired by \emph{SwarmLab} \cite{Soria2020}, a MATLAB-based simulator developed to create standardized processes and metrics for assessing the performance of swarm algorithms. 
SwarmLab functions as both a development tool and a comparative platform for the aerial swarm research community and for educational purposes. 
The simulator proposed in this study aligns with SwarmLab’s aim. 
Nevertheless, SPACE is distinct in that it is specifically tailored for MRTA research, whereas SwarmLab concentrates on the navigation of aerial swarms in cluttered environments. 
% SPACE highlights mission scenario generation and performance metrics that are specifically designed for MRTA. 
To minimize coding effort and support complex agent behaviors, the architecture of our simulator employs behavior trees and includes a graphical user interface tool for easy adjustment of these trees, enhancing both usability and flexibility in simulation setup.

\subsection{Multi-Robot Task Allocation}

% What is MRTA? (ST-MR, MT-SR)
As robotic systems mature, interest in deploying multiple robots has grown. 
This has brought attention to \emph{Multi-Robot Task Allocation} (MRTA) problems, which focus on determining which robot should perform which tasks to optimize system-level objective functions.
MRTA are classified into various taxonomies based on the characteristics of the robots and tasks involved (see \cite{Gerkey2004, Chakraa2023} for more details). 
Among these, the two most studied types are \emph{Multi-Task robots with Single-Robot tasks} (MT-SR) and \emph{Single-Task robots with Multi-Robot tasks} (ST-MR) \cite{Chakraa2023}.
To demonstrate the usefulness of the SPACE simulator, we compare representative decentralized algorithms for these two types (see Section \ref{sec_test}). 
Therefore, the remainder of this section provides brief reviews of these algorithms.

% CBBA (for MT-SR)
CBBA \cite{Choi2009} is a decentralized method for MT-SR scenarios, where each agent performs multiple tasks sequentially, and each task requires only one agent. 
Agents in CBBA greedily build task bundles by bidding on their desired tasks and sharing these bids with neighbors. 
Through a consensus process, they resolve inter-agent conflicts and determine task assignments without a central auctioneer. 
CBBA has inspired various extensions. 
ACBBA \cite{Johnson2011} introduces asynchronous operations, eliminating the need for agents to synchronize between the decision-making and conflict resolution phases.
CBBA with Partial Replanning (CBBA-PR) handles newly emerging tasks by resetting parts of previous allocations during bidding rounds \cite{Buckman2019}.
Grouped CBBA \cite{Kim2020} improves communication efficiency by organizing agents into groups.

% GRAPE (for ST-MR)
GRAPE \cite{Jang2018} is a decentralized algorithm that leverages an anonymous hedonic game framework. 
It was originally developed for ST-MR scenarios \cite{Jang2017, Hu2021}, where each agent is responsible for a single task that requires the collaboration of multiple robots. 
The algorithm aims to achieve a \emph{Nash stable partition}, where no agent has an incentive to unilaterally deviate, indicating that the swarm system has reached an agreed assignment.
A few GRAPE variants have recently emerged, extending to accommodate heterogeneous robots that can perform distinct services \cite{Diehl2023}. 
Some studies have enhanced convergence speed by utilizing a bipartite algorithm for initial partitioning \cite{Dutta2021} or by refining partition selection during its conflict-resolution process \cite{Wang2024}.

% Hungarian (ST-SR)

% Decentralised MRTA는 크게 2개의 로직으로 구성된다. 

% 사실 알고리즘 자체는 MT-SR, ST-MR 이렇게 나뉘어도, Sequential하게 하면 동일한 시나리오에서 다뤄볼수도 있다. 예를 들면 GRAPE을 sequential하게 하면 MT-SR문제를 풀수 있다. 그리고 그때의 결과를 CBBA와 비교해볼 수 있다. 우리가 제안하는 시뮬레이터는 이러한 알고리즘간의 비교 분석을 위해 제작되었다. 

% Dynmaic Task Scenario의 예시 {Park2021a}

\subsection{Contributions}

In this study, we propose SPACE, a Python-based simulator for MRTA research. 
We detail the development philosophy and software architecture behind its implementation (Section \ref{sec_architecture}). 
As a use case, we demonstrate how to implement CBBA and GRAPE as decision-making plugins within SPACE, specifically for scenarios where tasks are newly added over time. 
Additionally, we describe the adaptation of GRAPE, initially developed for ST-MR scenarios, to accommodate MT-SR scenarios (Section \ref{sec_implementation}).
We then provide a performance comparison of the two algorithms in terms of mission completion time, distance traveled by the agents, and tasks completed, discussing the distinct characteristics of each algorithm based on these metrics (Section \ref{sec_test}).

% Key Features of the Simulator
Below are the key features of the SPACE simulator:
\begin{itemize}
    \item \textbf{Swarm Robotics Focus}: Optimized for swarm robotics research, supporting large-scale simulations with a lightweight design.
    \item \textbf{Behavior Trees}: Uses behavior trees to define and manage agent behaviors, allowing for flexible and structured decision-making.
    \item \textbf{Groot2 Integration}: Compatible with \emph{Groot2}\footnote{\href{https://www.behaviortree.dev/groot/}{https://www.behaviortree.dev/groot/}} for visualizing and editing behavior trees, enhancing ease of use and analysis.

    \item \textbf{Flexible Configuration}: Easily configurable via a YAML file, enabling customization without modifying the source codes of the simulator.
    \item \textbf{Custom Plugins}: Integrates custom decision-making algorithms as plugins, allowing for tailored testing and experimentation.
    \item \textbf{Local Communication and Awareness}: Agents communicate locally within specified radii and maintain situational awareness based on their situational awareness ranges.
    \item \textbf{Dynamic Task Generation}: Supports the creation of tasks dynamically during the simulation, adapting to evolving scenarios.
    \item \textbf{Algorithm Comparison}: 
    Facilitates the comparison of different decision-making algorithms within a consistent simulation environment and supports Monte Carlo tests for statistical analysis.
\end{itemize}
The simulator is open-sourced\footnote{\href{https://github.com/inmo-jang/space-simulator}{https://github.com/inmo-jang/space-simulator}} and its tutorials are also available at the official documentation website\footnote{\href{https://space-simulator.rtfd.io/}{https://space-simulator.rtfd.io/}}.

% =================================================================================================
\section{Software Architecture}\label{sec_architecture}

\begin{figure*}[t]
\centering
\includegraphics[width=1.0
\textwidth]{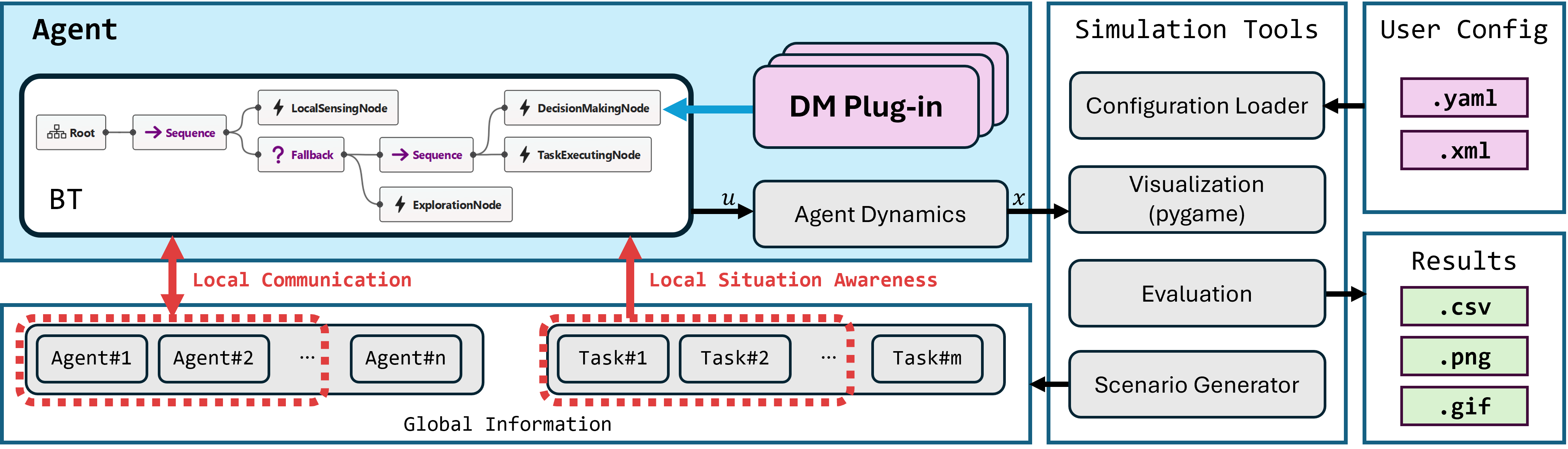}
\caption{SPACE Simulator Software Architecture}
\label{fig_overview}
\end{figure*}

The core components of SPACE, as shown in Figure \ref{fig_overview}, are as follows: behavior tree, agent, decision-making plug-in, task, and simulation tools.

\subsection{Behavior Tree}

% Importance of Behaviour tree
The simulator is developed using \texttt{pygame}\footnote{\href{https://www.pygame.org/}{https://www.pygame.org/}}, with each agent operating according to its own behavior tree. 
Behavior trees are increasingly popular in open-source robotics projects due to their modularity and flexibility \cite{Ghzouli2023}, which also motivated their use in our simulator. 
In each game loop iteration, the behavior tree of each agent is executed from the root. 
The simulator assumes that the behavior tree computation would be completed within a predefined control loop period, necessitating adjustments to the simulation frame rate parameter based on actual robot computation times for more realistic simulation. 

The default behavior tree is shown in Figure \ref{fig_overview}. 
During each loop, the agent starts with the \texttt{LocalSensingNode} to perceive any tasks nearby and receive any messages from its neighboring agents.
Using such local information, the agent executes the \texttt{DecisionMakingNode} for task assignments, where a decision-making plugin operates (see Section \ref{sec_decision_making_plugin}), and then moves to the \texttt{TaskExecutionNode} to carry out the assigned task.
If the decision-making process fails (e.g., if no tasks are perceived nearby), the agent executes the \texttt{ExplorationNode}, moving to a random position for a certain period to search for tasks.

% Why we implement our own behavior tree
We initially considered using an existing behavior tree library such as \texttt{py\_trees}; 
however, we opted to develop our own implementation to maintain a lightweight design for the simulator while ensuring greater flexibility for future enhancements. 
Notably, \texttt{py\_trees} lacks support for GUI tools \cite{Colledanchise2021}. 
By implementing our behavior tree in XML format, we made it compatible with \emph{Groot2}, a GUI tool designed for \texttt{BehaviorTree.CPP}, thereby allowing easy visualization and modification for behaviors of individual agents. 

% Current behaviour tree status
Our behavior tree implementation currently supports control nodes (e.g., \texttt{Sequence}, \texttt{Fallback}) and action nodes (e.g., 
\texttt{DecisionMaking} \texttt{Node}, \texttt{TaskExecutionNode},
\texttt{ExplorationNode}, \texttt{LocalSensingNode}).
Information exchange between action nodes is facilitated through a mechanism known as \emph{blackboard} \cite{Colledanchise2021, Ghzouli2023}. 
Users can extend functionality by adding custom action nodes, and a detailed tutorial for this is available on the official documentation website.
Custom behavior trees can be easily defined by dragging, dropping, and connecting nodes using Groot2.

% \begin{algorithm}
% \caption{Decision Making Plugin Example}
% \begin{algorithmic}[1]
% \State \textbf{Class} MyDecisionMakingPlugin
% \State \textbf{Methods:}

% \Function{init}{agent} 
%     \State self.agent $\gets$ agent
%     \State self.phase\_flag $\gets$ 1
    
% \EndFunction

% \Function{decide()}{}
%     \State Get information about neighbour tasks
%     \State Post-process if the previously assigned task is done

%     \If{self.phase\_flag = 1}
%         \State Do local decision-making process 
%         \State self.agent.message\_to\_share $\gets$ data to share
%         \State self.phase\_flag $\gets$ 2
%     \ElsIf{self.phase\_flag = 2}
%         % \State Do conflict-mitigating process using $\mathcal{M}_{rcv}$
%         \State Conflict-mitigating using self.agent.messages\_received
        
%         \State self.phase\_flag $\gets$ 1 if conflict-free 2 otherwise
%     \EndIf
% \EndFunction
% \end{algorithmic}
% \end{algorithm}

\subsection{Agent}\label{sec_agent}

The \emph{Agent} class encapsulates the core attributes and methods relevant for each agent in the simulator. 
Each agent instance possesses several fundamental attributes such as its status information (e.g.,  identification number, {position}, {velocity}, {acceleration}, {rotation}), and its mobility capabilities (e.g., maximum linear speed, maximum angular speed, maximum acceleration). 
Each agent instance also has an attribute called work rate, which specifies the amount of task workload the agent can perform per second.

% Local Communication Capability    
Each agent also has attributes related to its local perception capabilities. 
These include its communication range, which defines the distance within which the agent can interact with neighboring agents, and its situational awareness range, which determines the area within which the agent can perceive nearby tasks.
A crucial function for local communication, \texttt{local\_message\_receive()}, collects local decision data from neighboring agents. 
When each agent executes its decision-making process at the \texttt{DecisionMakingNode}, it stores the resulting local decision data in \texttt{message\_to\_share}, a Python dictionary.
The use of a dictionary allows users to define the message structure flexibly within their decision-making plugins (see Section \ref{sec_decision_making_plugin}), adapting the format as needed. 
The \texttt{local\_message\_receive()} function  retrieves these local data and appends them to a Python list \texttt{messages\_received} of the agent. 
This function is invoked by the \texttt{LocalSensingNode} of the behavior tree in each game loop.

Agents are modeled as point masses and move in straight lines toward their assigned tasks, which is executed at the \texttt{TaskExecutingNode}.
Although practical implementations involve obstacle avoidance and inter-agent collision avoidance, the simulator omits these complexities for a clearer focus on evaluating high-level decision-making algorithms, which are the primary subject of the analysis.

\subsection{Decision-making Plug-in}\label{sec_decision_making_plugin}

Inspired by the approach used in \emph{Navigation2} \cite{Macenski2020}, which allows for easy replacement of planners or controllers, our simulator is designed to enable users to effortlessly swap decision-making algorithms by modifying \texttt{config.yaml} file, treating it as a plugin. 
Three decision-making plugins are currently implemented and will be detailed in Section \ref{sec_implementation}.

Users can also implement their own decision-making algorithms as custom plugins in separate Python files. 
Our simulator aims to simplify the process of implementing algorithms from pseudocodes found in the literature. 
% Our simulator aims to simplify this process by enabling users to implement algorithms directly according to the pseudocodes described in the literature, reducing the need for additional coding. 
For instance, decentralized algorithms typically involve a two-step process \cite{Chopra2017, Choi2009, Jang2018}: first, each agent performs local decision-making and shares the results with neighboring agents; second, the agent uses the shared information to resolve conflicts.
Handling local communication requires extra coding efforts, but the SPACE simulator streamlines this process by providing built-in support for local communication.
Users can implement the first step by simply storing data to be shared (e.g., local decision data) in the \texttt{message\_to\_share} attribute of an agent before exiting the \texttt{DecisionMakingNode}. 
The simulator automatically manages the distribution of this data, ensuring that, in the next game loop, each agent receives the data from its neighboring agents during the \texttt{LocalSensingNode}. 
Consequently, agents can then use this data to perform conflict resolution.

\subsection{Task}
The \emph{Task} class contains attributes for each task, including its identification number, position, and workload.
To manage these tasks, two key functions are utilized. 
The \texttt{reduce\_amount()} function decreases the workload according to the work rate of an agent performing the task.
The \texttt{set\_done()} function is called when the workload reaches zero, updating the \texttt{completed} attribute of the task to \texttt{True}. 
This status is then made available to neighboring agents whose situational awareness range includes the task.

% Dynamic Task Generation Scenario
Task instances are initially generated at the start of the simulation in a specified number within a defined area, using a uniform random distribution. 
After this initial setup, additional tasks can be created periodically at defined time intervals. 
The frequency and number of the additional tasks can be configured through \texttt{config.yaml} file. 
This dynamic task generation feature allows for the evaluation of decision-making algorithms in terms of how effectively they can adapt to changes in the environment.

\subsection{Other Features}\label{sec_other_features}

Upon completion of a simulation, metrics such as the distance traveled and workload completed by the agents are recorded, not only offering a time-wise summary of the simulation progress but also capturing the individual agent-level data at the end of the mission. 
Both results are saved in CSV files. 

The simulator supports two testing modes: with screen rendering and without. In screen rendering mode, the simulation can be visualized using \texttt{pygame}, and the output can be saved as a GIF through an optional feature.

A notable feature of this simulator is its support for Monte Carlo experiments. 
To conduct these experiments, users begin by configuring a scenario in a \texttt{config.yaml}, specifying parameters such as the decision-making plugin, agent, and task settings. 
Multiple configurations can be developed to explore various parametric studies. 
A set of these configurations can be defined in a separate YAML file, where the number of Monte Carlo runs for each scenario is also specified. 
The simulator then automatically performs the tests repeatedly according to the defined parameters.

% =================================================================================================
\section{Plug-in Implementation Examples}\label{sec_implementation}

This section presents examples of how to implement decision-making plugins within the SPACE simulator. 
We implemented \emph{CBBA} \cite{Choi2009}, \emph{GRAPE} \cite{Jang2018}, and a simple algorithm named \emph{First-Claimed Greedy}. 
Drawing from our development experience, we outline a recommended structure for decision-making plugins. 
We then discuss additional modifications made to GRAPE and CBBA, which extend beyond their original descriptions to support the MT-SR type dynamic task generation scenario in Section \ref{sec_test}.

\subsection{Structure Overview}

Basically, a decision-making plugin is supposed to be defined as a Python class (see Class \ref{algorithm_dm_plugin}), with a member function \textsc{decide()} that is invoked by the \texttt{DecisionMakingNode} in the
behavior tree. 
The node acts as a wrapper that connects the behavior tree with the plugin.
Any elements that require initialization at the start of the mission should be placed in a separate initialization function. 
For example, the boolean variable \textsf{\small satisfied} (Line 7) should be initially set to \texttt{False} (not shown here for simplicity).

% GRAPE Implementation
\begin{algorithm}
\caption{DM Plug-in Structure}
\begin{algorithmic}[1]

\Function{decide()}{}
    \State Get \textsf{\small local\_tasks} and \textsf{\small local\_tasks} from \textsf{\small blackboard} \label{line_local_sensing}
    \State Post-process if the previously assigned task is done \label{line_assigned_task_postprocessing}
    % Get information about neighbour tasks
    \If{no task in \textsf{\small local\_tasks}}
        \State \textbf{return} None
    \EndIf

    \If{ \textsf{\small satisfied} = \texttt{False}}  \label{line_plugin_main_start}       
        \State Local decision-making \label{line_local_decision_making}
        
        % \State Algorithm 1 (L5-10)  in \cite{Jang2018} \Comment{Decision-making}
        
        \State \textsf{\small message\_to\_share}  $\gets$ data to share \label{line_message_to_share}
        \State \textsf{\small satisfied} $\gets$ \texttt{True} 
        \State \textbf{return} \texttt{None} \label{line_satisfied_true}
    \Else
        % \State Do conflict-mitigating process using $\mathcal{M}_{rcv}$    
        
        \State Conflict-mitigating using \textsf{\small messages\_received}  \label{line_conflict_mitigating} 
        
        % \State Algorithm 2 in \cite{Jang2018} \Comment{Conflict-mitigating}
        % \Statex \quad\quad\quad using \textsf{\small messages\_received}  
        
        \State \textsf{\small satisfied} $\gets$ \texttt{False} if not conflict-free yet \label{line_conflict_not_free_yet}
        
        % \State \textsf{\small satisfied} $\gets$ \texttt{True} if conflict-free 
        % \Statex \quad\quad\quad\quad\quad\quad\quad\quad \texttt{False} otherwise

        \State \textbf{return} \textsf{\small assigned\_task\_id} if \textsf{\small satisfied} = \texttt{True} \label{line_output}
        \Statex \quad\quad\quad\quad\quad\quad \texttt{None} otherwise        
    \EndIf \label{line_plugin_main_end}
\EndFunction
\end{algorithmic}\label{algorithm_dm_plugin}
\end{algorithm}

The decision-making plugin starts by retrieving local information from the blackboard (Line \ref{line_local_sensing}), which was gathered by the \texttt{LocalSensingNode} at the beginning of the behavior tree, as shown in Figure \ref{fig_overview}.
If there is already an assigned task, the plugin checks its completion status and takes necessary post-completion actions (Line \ref{line_assigned_task_postprocessing}). 
For instance, in CBBA, this involves removing the completed task from the task bundle and designating the next task in the list as the new assignment.
If no local tasks are found on the blackboard, the decision-making process concludes with \texttt{None} (Lines 4-6), causing the \texttt{DecisionMakingNode} to signal a failure and triggering the \texttt{Fallback} to activate the \texttt{ExplorationNode} to search for other tasks in the environment.

Decentralized MRTA algorithms typically involve a \emph{local decision-making} process followed by a \emph{conflict-mitigating} process using information from neighboring agents. 
To distinguish between these two steps within the  plugin, we use a boolean flag, \textsf{\small satisfied}. 
When \textsf{\small satisfied} is \texttt{False}, the decision-making process is executed as required (Line \ref{line_local_decision_making}). 
For GRAPE, this corresponds to Algorithm 1 in \cite{Jang2018}, and for CBBA, Algorithm 3 in \cite{Choi2009}. 
Next, any messages to share with neighboring agents are stored in \texttt{message\_to\_share}, and \textsf{\small satisfied} is set to \texttt{True} (indicating the agent is content with its decision), after which the function ends (Lines \ref{line_message_to_share}-\ref{line_satisfied_true}). 

In the next game loop, the \texttt{message\_received} will contain messages from its neighboring agents. 
Since \textsf{\small satisfied} was set to \texttt{True} in the previous loop, the \texttt{DecisionMakingNode} skips the decision-making process and instead runs the conflict-mitigating process (Line \ref{line_conflict_mitigating}). 
For GRAPE, this is Algorithm 2 in \cite{Jang2018}, and for CBBA, it is the implementation of Table 1 in \cite{Choi2009}. 
If the conflict is resolved, \textsf{\small satisfied} remains \texttt{True}; otherwise, it is set back to \texttt{False} (Line \ref{line_conflict_not_free_yet}), prompting the decision-making process in the subsequent game loop. 
When the conflict is resolved for every agent, it means that GRAPE has identified a Nash stable partition or that CBBA has finalized the task bundle list.
Based on this converged information, each agent can determine its assigned task and returns the task identification number as output (Line \ref{line_output}).
Then, the \texttt{DecisionMakingNode} records the identification number in the blackboard, allowing the \texttt{TaskExecutionNode} to direct the agent toward the task.

To summarize, if users follow the proposed structure, they only need to implement Lines \ref{line_local_decision_making} and \ref{line_conflict_mitigating} as specified in the pseudocodes of the literature, minimizing the need for additional modifications. 
A template for this is available in the SPACE official documentation website.

\subsection{GRAPE Modifications}

In the scenario involving dynamic task generation, additional modifications are made to the original GRAPE. 
First, we incorporate a mechanism where each agent decentralizedly constructs an initial partition such that each neighbor agent is assigned a task closest to it as the starting point. 
As shown by \cite{Dutta2021}, setting an initial partition can accelerate convergence to a Nash stable partition, thereby facilitating better adaptation to dynamic environments.
This process is implemented at the initialization function of the decision-making plugin and also during the post-processing after a task is completed (Line \ref{line_assigned_task_postprocessing}).

Although GRAPE has been theoretically proven to converge to a Nash stable partition when agents are connected in a strongly connected topology, determining whether all agents have converged requires global information. 
Hence, we implement such that each agent assesses convergence based solely on its local information.
Specifically, the agent compares its locally-known partition with the partition resulted from the distributed-mutex algorithm \cite{Jang2018} (Line \ref{line_conflict_mitigating}). 
If there is no difference between the two partitions, the agent concludes that its locally-known partition becomes Nash-stable, and \textsf{\small satisfied} remains \texttt{True} (Line \ref{line_conflict_not_free_yet}).
However, if the agent receives a new partition information from another agent further than one hop away,  
\textsf{\small satisfied} might be set to \texttt{False}.  
This triggers to proceed the decision-making process based on the new information.

For the individual utility function of agent $a_i \in \mathcal{A}$ with respect to $t_j$, we use the following equation:
\begin{equation}\label{eqn_utility_ftn_GRAPE}
    u_i(t_j, |S_j|) = \frac{R_j}{|S_j|} - c_{i,j} \cdot {|S_j|}^{f_s},
\end{equation}
where $S_j \subseteq \mathcal{A}$ is the task-specific coalition; 
$R_j$ is the reward for completing task $t_j$; 
$c_{i,j}$ is the distance-based cost from agent $a_i$ to $t_j$; 
and $f_s \in \mathbb{R}^+$ is a scalar factor representing social inhibition between agents. 
To adapt GRAPE, originally designed for ST-MR scenarios, to MT-SR scenarios, we introduce the term $|S_j|^{f_s}$. 
When the coalition size is 1, this term has no effect, making Equation (\ref{eqn_utility_ftn_GRAPE}) identical to the form in \cite{Jang2018}. 
However, as the coalition size increases, this term amplifies the cost, encouraging agents to disperse, which facilitates handling MT-SR scenarios.

\subsection{CBBA Modifications}

In dynamic environments, as noted by \cite{Johnson2011, Buckman2019}, CBBA may need to be rerun to address outdated information or significant changes in situational awareness. 
Our empirical experiments also reveal that discrepancies between bid costs and actual execution results further affect the phenomenon.
For instance, we observed scenarios where an agent, after winning a task with a high bid value, later removed the task from its task bundle when a more attractive one emerged.
Consequently, the task remained unassigned, as the other agents were unable to bid on it due to its high cost.

To address the problem of invalid bid values and unassigned tasks, we introduce a simple mechanism: if the task bundle of an agent remains empty for a certain period, the agent resets all of its known winning bid values and winning agent IDs. 
Although resetting bid values was also proposed in CBBA-PR \cite{Buckman2019}, our study aims to compare the baseline versions of CBBA and GRAPE. 
Thus, we implement this simple modification to CBBA instead of using CBBA-PR to evaluate its effectiveness in dynamic environments.

For the scoring function of agent $a_i$ with regard to its task-execution path $\mathbf{p}_i$, we use the \emph{time-discounted reward} \cite{Choi2009}:  
\begin{equation}\label{eqn_utility_ftn_CBBA}
    S_i^{\mathbf{p}_i} = \sum \lambda^{ \tau_i^j(\mathbf{p}_i)} \cdot R_j,
\end{equation}
where $\lambda \in [0, 1]$ is the discount factor;   $\tau^j_i \in \mathbb{R}^+$ represents the estimated time for the agent to finish task $t_j$, taking into account the distance to the task along the path, and its work rate. 
$R_j$ is the reward received upon task completion.  

\subsection{First-Claimed Greedy Algorithm}

We implement a simple algorithm where each agent selects the task nearest to itself based on local task information. 
In making this selection, the agent also considers messages received from its neighboring agents. 
If the agent realizes that the chosen task has already been assigned to another agent, it abandons that task and tries the task selection process again in the next game loop.
We refer to this approach as the \emph{First-Claimed Greedy} algorithm, as it ensures that the initial claim of an agent on a task is confirmed. 
% For its implementation, Lines \ref{line_plugin_main_start}-\ref{line_plugin_main_end} in Class \ref{algorithm_dm_plugin} are simplified into  Lines \ref{line_conflict_mitigating} to \ref{line_output}. 
% The core logic described above is succinctly implemented in Line \ref{line_conflict_mitigating}.

% =================================================================================================
\section{Use Case: Comparison of Algorithms}\label{sec_test} 

This section demonstrates the usefulness of the SPACE simulator by comparing three algorithms, GRAPE, CBBA, and First-Claimed Greedy (FCG), in a MT-SR scenario. 
Our goal aims to showcase how SPACE can be used to evaluate and contrast these algorithms effectively.

In the scenario, each agent can sequentially execute multiple tasks, with each task requiring at least one robot for completion. 
While multiple robots may collaborate on a single task if necessary, this can increase the associated costs for those robots. 
When such collaboration occurs, the work rates of the participating agents are combined, resulting in faster task completion. 
Although in practice the summation of work rates might exhibit submodular characteristics, for simplicity in this evaluation, we assume a linear summation.

\subsection{Experiment Setting}

% 본 섹션에서는 MT-SR 시나리오에 GRAPE와 CBBA, 그리고 First-Claimed Greedy (FCG) Algorithm을 적용한 비교분석 결과를, 본 SPACE simulator 활용의 use case로서 보여주도록 하겠다.

\begin{table}[h!]
\caption{Overview of Experimental Parameters}\label{table_experiment_parameters}
\centering
\begin{tabular}{|l|l|}
\hline
\textbf{Feature} & \textbf{Parameter Value} \\
\hhline{|=|=|}
Simulation Update Rate & 1 Hz \\
\hline
Area Size ($w, h$) & 1400, 1000 \\
\hline
Number of Tasks ($n_t$) & 250 (initial); 50 $\times$ 3 (additional) \\
\hline
Task Workload & [6, 60] \\
\hhline{|=|=|}
Agent Work Rate & 1 \\
\hline
Max Linear Speed ($v_{max}$) & 0.25 \\
\hline
Max Acceleration ($a_{max}$) & 0.01 \\
\hline
Max Angular Speed ($\omega_{max}$) & 0.25 \\
\hline
Situation Awareness Range ($r_p$) & 300 \\
\hline
Communication Range ($r_c$) & \{100, 200, 300\} \\
\hline
Number of Agents ($n_a$) & \{10, 30, 50\} \\
\hline
MRTA Algorithm & \{CBBA, GRAPE, FCG\} \\
\hline
\end{tabular}
\end{table}

The common settings for the scenario are as follows. 
Basically, the simulation loop occurs with 1 Hz with consideration of potentially low communication bandwidth between agents. 
The map has dimensions of 1400 units in width and 1000 units in height. 
The scenario starts with 250 tasks, with an additional 50 tasks being generated every 1000 seconds for 3 times, resulting in 400 tasks in total. 
These tasks are distributed randomly across the map according to a uniform distribution. 
Each task takes between 6 and 60 seconds to complete, as all agents work at the same work rate of 1.
Each agent has the same mobility capabilities as shown in Table \ref{table_experiment_parameters}, and has a situation awareness range with 300 units. Regarding the utility functions for GRAPE and CBBA, i.e., Equations (\ref{eqn_utility_ftn_GRAPE}) and (\ref{eqn_utility_ftn_CBBA}), we use the following parameters: $R_j$ is set to be the workload of task $t_j$; the social inhibition factor $f_s$ is 100; the discount factor $\lambda$ is 0.999. 

Our experiments evaluate the performance of GRAPE, CBBA, and FCG algorithms under varying conditions. 
We vary the number of agents $n_a \in \{10, 30, 50\}$ and communication radii $r_c \in \{100, 200, 300\}$ to assess their impact on each algorithm.
Specifically, we test a total of 27 scenarios, corresponding to all combinations of agent numbers, communication radii, and algorithms, with 100 Monte Carlo simulations conducted for each scenario.
The performance of each algorithm is assessed based on three key metrics: (1) mission completion time, (2) distance traveled by the agents, and (3) workload done by the agents. 
Note that all experiments are conducted using SPACE's Monte Carlo simulation support, as detailed in Section \ref{sec_other_features}.

\subsection{Comparative Results}

\begin{figure*}[t]
\centering
\subfigure[$n_a = 10$]{\includegraphics[width=0.30\linewidth]{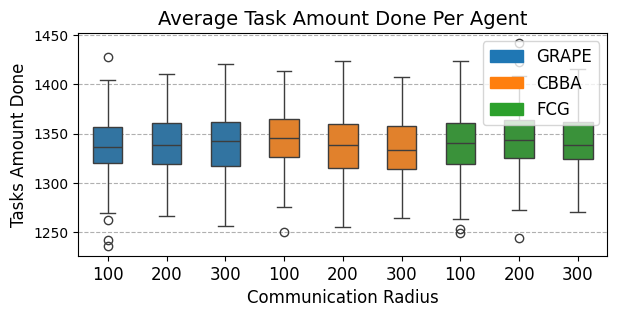}}
\hfil
\subfigure[$n_a = 30$]{\includegraphics[width=0.30\linewidth]{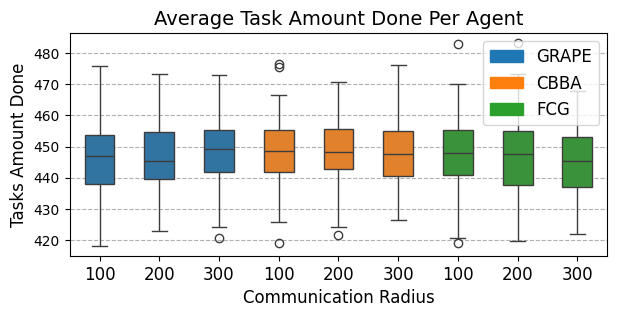}}
\hfil
\subfigure[$n_a = 50$]{\includegraphics[width=0.30\linewidth]{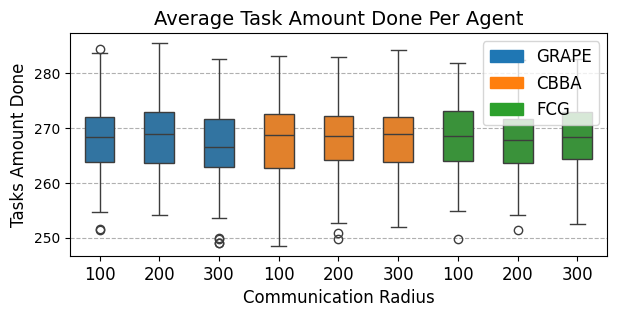}}
\vfill
\subfigure[$n_a = 10$]{\includegraphics[width=0.30\linewidth]{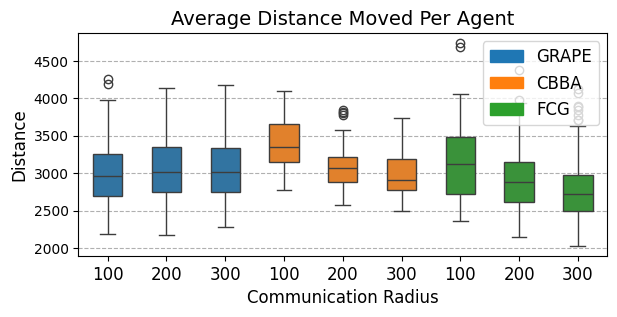}}
\hfil
\subfigure[$n_a = 30$]{\includegraphics[width=0.30\linewidth]{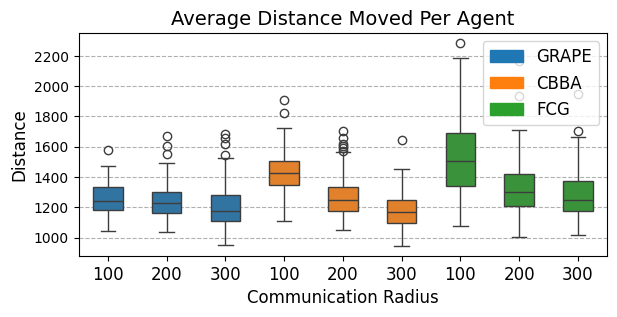}}
\hfil
\subfigure[$n_a = 50$]{\includegraphics[width=0.30\linewidth]{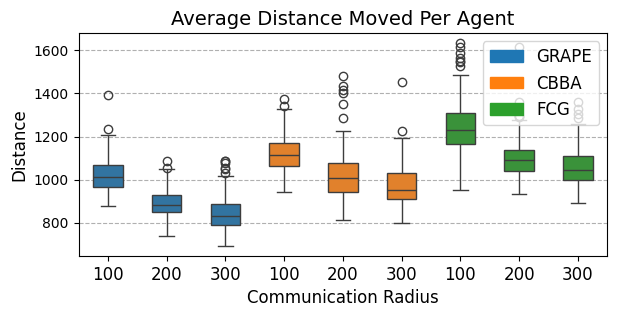}}
\vfill
\subfigure[$n_a = 10$]{\includegraphics[width=0.30\linewidth]{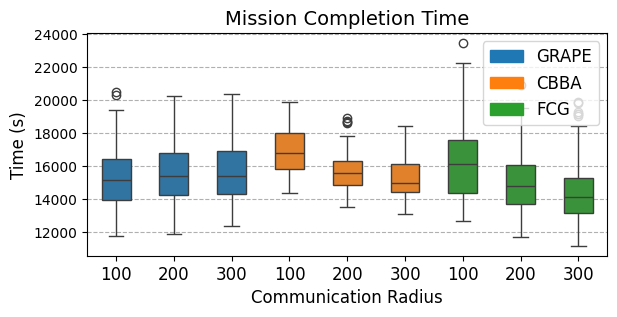}}
\hfil
\subfigure[$n_a = 30$]{\includegraphics[width=0.30\linewidth]{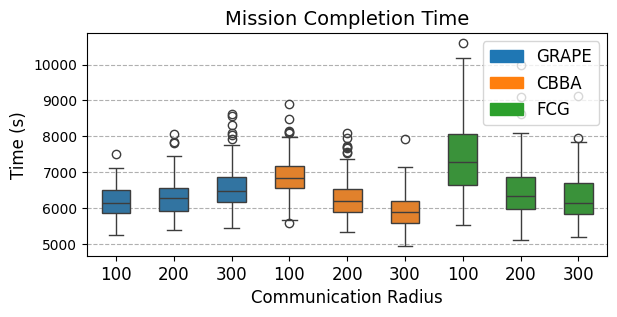}}
\hfil
\subfigure[$n_a = 50$]{\includegraphics[width=0.30\linewidth]{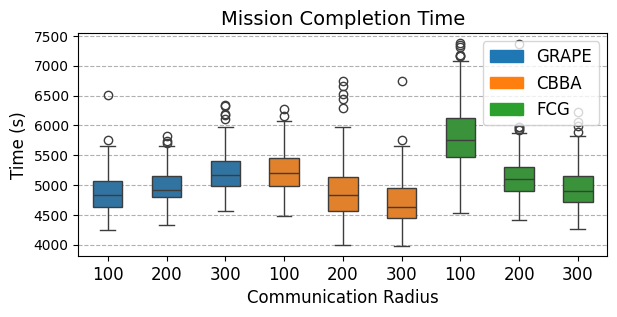}}
\caption{Comparative performance results of GRAPE \cite{Jang2018}, CBBA \cite{Choi2009}, FCG in scenarios with $n_a \in \{10, 30, 50\}$ and $r_c \in \{100, 200, 300\}$.}
\label{fig_result_performance}
\end{figure*}

Figure \ref{fig_result_performance} presents the results of mission completion time, average travel distance per agent, and average workload per agent for scenarios with 10, 30, and 50 agents over 100 episodes. 
These results are visualized using boxplots, with varying communication ranges of 100, 200, and 300, and for the algorithms GRAPE (blue), CBBA (orange), and FCG (green).
% Figure \ref{fig_result_performance} the number of agents가 10, 30, 50인 경우에 대해서, mission completion time에 대한 결과; 개별 agent의 평균 이동거리; 개별 agent의 평균 workload done를 100개의 episodes에 대해서 boxplot을 그린 것이다.  
% 한 그래프에 GRAPE, CBBA, FCG에 대해서, communication range를 100, 200, 300으로 변화시킨 결과을 가시화하였다. GRAPE는 파랑색으로, CBBA는 오렌지색, FCG는 녹색으로 표현하였다. 

\subsubsection{Average Workload Per Agent}
Figure \ref{fig_result_performance}(a)-(c) visualizes the average workload per agent for each scenario. 
Notably, regardless of the number of agents, algorithm used, or communication range, the average workload per agent remains consistent. 
This can be explained by the fixed number of tasks. 
Given that the task workload ranges between [6, 60] with a total of 400 tasks, the expected workload sum is approximately 13,200. 
Thus, with the fixed number of tasks, increasing the number of agents leads to a decrease in individual agent workload. 
Since all tasks must be completed to conclude the mission, the average workload per agent remains unaffected by variations in communication range or the choice of algorithm.
% 각 결과에 대해서 해석이 쉬운 것 부터 설명을 해보려고 한다. 
% 우선, Figure \ref{fig_result_performance}(g)-(i)는 각 시나리오에서의 agent당 평균 workload를 보여주는데, 동일한 agents숫자 일때, 알고리즘이 무엇이든, communication range가 어떻게 되든 상관없이 동일한 값을 보이고 있다.
% 일단 해당 값의 수준의 적절성을 살펴보면, Task workload의 random generation 범위가 [6,60]이고 개수가 400개라는 점을 고려하면 13200이 예상값이고, 이것을 고려하여 각시나리오에 involved된 agents숫자를 고려하면, 이러한 결과값이 나오는 것은 당연하다. 
% 그래서 고정된 tasks 숫자이기 때문에, agents 숫자가 많아질 수록 개별 agent의 workload 수행량은 줄어든다. 
% communication range나 알고리즘에 영향을 받지 않는 이유는, 어차피 시나리오상 주어진 tasks 총량은 동일하고, 이것이 어쨌든 모두 수행되어야 mission이 종료가 되기 때문에, agent별 평균 작업량 자체는 알고리즘이나 communication range에 영향을 받지는 않는다.

\subsubsection{Average Travel Distance Per Agent}
Figure \ref{fig_result_performance}(d)-(f) illustrates the average travel distance per agent. 
As communication range increases, the travel distance decreases for all algorithms and scenarios. 
This effect is more noticeable with a higher number of agents. 
A shorter communication range leads to conflicts between agents that are not immediately aware of each other during local decision-making processes. 
When the agents eventually become aware of these conflicts, they often resolve the conflicts by rerouting, which increases their travel distance.
This is particularly evident in FCG, where the agents do not abandon their assigned tasks even upon detecting other agents heading to the same task. 
Hence, shorter communication ranges result in increased average distances due to more frequent rerouting.
% 두번째로 Figure \ref{fig_result_performance}(d)-(f)는 각 시나리오에서의 agent당 평균 이동거리를 보여주는데, 이것은 모든 알고리즘 및 모든 시나리오 에서 communication range가 커질수록 이동거리가 짧아지는 모습을 보인다. 특히 agents 숫자가 커질 수록 communication range에 의한 영향이 모든 알고리즘에서 두드러지게 나타난다. 그럴 듯한 이유중 하나는, communication range가 짧을 수록 local decision을 수행하고 해당 task로 이동하는 과정에서, conflict가 있는 다른 agents (예를 들어, decision-making을 수행할 때는 멀리 떨어져있어서 몰랐으나, task를 향해 서로 가까이 이동하면서 서로의 decision 정보를 알게 되었을 때)를 알게 되었을 때, 각 알고리즘의 conflict mitigating process에 따라서, 둘 중 하나가 다른 task로 이동하게 되어, 이동거리가 늘어난다. 즉, communication range가 좁을 수록 이러한 불필요한 이동이 늘어나고, average distance가 늘게 된다. 
% 이러한 효과는 특히 FCG에서 더 나타나는데 이는 FCG의 경우 한번 local decision을 결정하면, 동일한 task로 향하는 다른 agent를 인지하게 되더라도 한 agent가 포기하는 로직이 없기 때문이다. 

\subsubsection{Mission Completion Time}
Figure \ref{fig_result_performance}(a)-(c) shows mission completion time. 
Interestingly, the impact of communication range on mission completion time varies by algorithm. 
For CBBA and FCG, mission completion time decreases as communication range increases. 
In CBBA, each agent only needs to verify the convergence of its task bundle. 
Even if some other agents within the communication radius have not yet converged on their bundles, conflicts are avoided as long as the agent does not prefer any of the tasks selected by them. 
For each agent, once its task bundle converges, it proceeds to execute the assigned tasks. 
 
However, in the case of GRAPE, since the partition information itself must converge, even if other agents select tasks without conflict, the partition information is still affected, resulting in \texttt{False} after the distributed mutex algorithm. 
This effect extends the convergence time as the communication range widens.

% 마지막으로 Figure \ref{fig_result_performance}(a)-(c)는 mission completion time을 나타낸다. 이것은 흥미롭게도 communication range가 decision-making에 미치는 영향이 알고리즘에 따라 다름을 보여준다.
% communication range가 클수록 CBBA와 FCG는 mission completion time이 줄어드는 추세를 보이는데, GRAPE는 반대로 늘어나는 추세를 보인다. 
% communication range가 넓을 수록 더 좋은 결과를 보이는 것이 직관적인 해석이다. 특히 Figure \ref{fig_result_performance}(d)--(f)와 (g)--(i)를 보면, 모든 알고리즘에서, communication ragne가 커짐에 따라 agents의 distance traveled이 줄어들고, task workload done은 유지가 되었다. 따라서 communication range가 넓을 수록 더 좋은 결과를 보이는 것이 직관적인 해석이다. 
% 하지만 GRAPE의 경우, 그럼에도 불구하고 communication range가 커짐에 따라 mission completion time이 커진다. 
% 이는 GRAPE의 수렴 과정의 특징으로 이것에 소요되는 시간이 늘어나기 때문이다. 구체적으로 말하자면, 
% CBBA는 각 agent는 자신의 task bundle이 수렴되었는지만 체크하면 되기 때문에, comm radius내에 다른 agent가 bundle 수렴이 안되더라도, 자기가 선택한 task를 선호하지 않는 다면 해당 agent와 conflict가 일어날 일이 없다. 따라서 개별적으로는 수렴이 되었기 때문에 그런 agents는 assigned tasks로 작업을 위해 이동을 진행하게 된다. 오히려 comm radius가 넓을 수록 한번에 winning bid 및 winning agents 체크가 빠르게 되기 때문에, comm radius가 넓은 것이 수렴에 효과가 있다. 

% 하지만 GRAPE의 경우 partition information 자체가 수렴이 되어야 하기 때문에, 다른 agent가 conflict가 없는 task를 선택한다고 하더라도 partition 정보 자체에 영향을 주기 때문에 the distributed mutex algorithm을 거치고 나면 satisfied가 False가 되게 된다. 이러한 효과는 communication range가 넓어질 수록 수렴시간을 더 늘리게 된다. 
% 하지만 그럼에도 흥미로운 점은 GRAPE은 communication range를 100으로 하는 것이 다른 알고리즘의 best case와 어느 정도 비슷한 결과를 보인다는 점이다.

\subsubsection{Comparison of CBBA and GRAPE}

The results with 50 agents highlight the most notable trends. 
CBBA with $r_c = 300$ achieves the best mission completion time (Figure \ref{fig_result_performance}(i)), while GRAPE results in the shortest average distance  (Figure \ref{fig_result_performance}(f)). 
These findings suggest that CBBA may be more suitable for scenarios requiring faster completion, whereas GRAPE could be advantageous for missions where minimizing travel distance is critical. 
However, it is important to note that these outcomes are based on the specific settings of this study, and therefore, should be interpreted with caution before making broader generalizations.

% 가장 경향성이 두드러지는 agents 수가 50일 때를 기준으로 알고리즘 별 비교를 해보면, mission completion time 측면에서 가장 best는 CBBA가 communication range 300을 했을 때다 (Figure \ref{}(i)). 
% 하지만 average distance 측면에서는 GRAPE가 communication range 300으로 했을 때다. 이는 본 실험셋팅에서의 결과상, 필요에 따라서 빠른 수행은 CBBA가 적합할 수 있고, 이동 거리가 중요한 미션에서는 GRAPE이 적절할 수 있음을 보여준다. 하지만 이는 구체적으로 agent utility function or scoring scheme을 어떻게 하느냐에 따라서도 영향을 줄수 있다. 

% =================================================================================================
\section{Conclusion}

In this study, we proposed SPACE, a Python-based simulator designed for MRTA research, and outlined the design philosophy and software architecture underlying its development. 
To demonstrate its practical utility, we implemented CBBA and GRAPE as decision-making plugins within SPACE and compared their performance across various metrics, particularly in scenarios where tasks are dynamically introduced over time.
This evaluation revealed distinct characteristics of each algorithm, providing valuable insights into their respective strengths and weaknesses. 
These findings highlight the usefulness of SPACE as a tool for conducting in-depth comparisons and analyses of different algorithms, thereby facilitating future research in MRTA.

For future work, we plan to enhance SPACE by developing a reinforcement learning interface that functions similarly to the OpenAI Gym environment.
This enhancement will enable SPACE to support multi-agent reinforcement learning \cite{Na2023} specifically focused on MRTA research. 
Additionally, we aim to broaden the range of scenarios beyond the currently implemented MT-SR scenario, including more complex scenarios such as delivery task allocation \cite{Chen2021}, where tasks have constraints on starting and ending positions.
Extending the behavior tree by adding behavior nodes for path planning and control could be a valuable direction as well.

\addtolength{\textheight}{-12cm}   % This command serves to balance the column lengths

%%%%%%%%%%%%%%%%%%%%%%%%%%%%%%%%%%%%%%%%%%%%%%%%%%%%%%%%%%%%%%%%%%%%%%%%%%%%%%%%

% \section*{ACKNOWLEDGMENT}

% ......

%%%%%%%%%%%%%%%%%%%%%%%%%%%%%%%%%%%%%%%%%%%%%%%%%%%%%%%%%%%%%%%%%%%%%%%%%%%%%%%%

%\bibliographystyle{ieee_fullname}
\bibliographystyle{ieeetran}
\bibliography{library.bib}

\end{document}